\documentclass[11pt]{article}
\usepackage{xcolor}

\usepackage[final]{acl}

\usepackage{times}
\usepackage{latexsym}
\usepackage{booktabs}
\usepackage{float}

\usepackage[T1]{fontenc}

\usepackage[utf8]{inputenc}

\usepackage{microtype}

\usepackage{inconsolata}

\usepackage{graphicx}

\usepackage{listings}
\usepackage{makecell}
\usepackage{array}

\usepackage{booktabs}
\usepackage{tabularx}
\usepackage{enumitem}

\title{TurEngMix: A Text Corpus and Benchmark for Turkish-English Code-Mixed Language Identification and Named Entity Recognition}

\author{
  Ilayda Dogan\\
  Computer Science \\
  University of Maryland \\
  \texttt{idogan@umd.edu}
  \And
  Phuong-Anh Nguyen-Le \\
  Information Science \\
  University of Maryland\\
  \texttt{nlpa@umd.edu}
  \And
  Julia Mendelsohn \\
  Information Science \\
  University of Maryland\\
  \texttt{juliame@umd.edu}
}

\begin{document}
\maketitle
\begin{abstract}

Natural language processing systems underperform on code-mixed text, particularly for low-resource language pairs. Turkish-English poses a further challenge: it lets English stems combine with Turkish suffixes to form single mixed-language tokens. We introduce TurEngMix, a corpus of 5.5K noisy, naturally occurring social media posts (486,974 tokens) rich in Turkish-English code-mixing. From this corpus, we construct a new Turkish-English benchmark for code-mixed language identification (LID) and named entity recognition (NER), comprising 15K expert-annotated tokens. Evaluating both decoder LLM and fine-tuned encoder baselines, we find that monolingual Turkish and English tokens are labeled reliably, but all models have high error rates on mixed-language tokens for both LID and NER. For morphologically integrated tokens, NER error rates were 5.2× and 6.3× higher for GPT-4o and Qwen, respectively. This highlights how morphological integration remains a challenge.
We release the corpus, annotations, and code to support future computational and sociolinguistic research on Turkish-English code-mixing. 

\end{abstract}

\section{Introduction}

Code-mixing, the use of two or more languages within a single sentence or phrase, is widespread in online spaces and diaspora communities \cite{poplack-1980, androutsopoulos2013code}. It serves a range of linguistic and social functions, including identity signaling, tone-setting, and communicating culturally or technically specific concepts, and takes different structural forms across speaker backgrounds and communicative contexts \cite{tay1989code}. However, despite its natural role in multilingual communities and growing presence in informal, online settings, code-mixing remains underexplored in natural language processing (NLP), particularly for lower-resourced language pairs \cite{winata2023decades}, reflecting the broader language diversity gap in NLP \cite{joshi-etal-2020-state, pakray2025natural}.

Existing code-mixing benchmarks are dominated by language pairs such as Spanish-English and Hindi-English \cite{dogruoz-etal-2021-survey}, where code-mixing typically occurs at word boundaries. Turkish, however, presents a markedly different linguistic setting. Its agglutinative morphology allows suffixes expressing case, possession, tense, and other grammatical functions to stack into complex words. In Turkish-English code-mixed discourse, English lexical stems are productively integrated into this structure \cite{yirmibesoglu-eryigit-2018-detecting}, resulting in
\textit{mixed-language tokens} that combine English roots with Turkish suffixes (examples in Table~\ref{tab:examples-intro}). These naturally occurring forms are linguistically interesting manifestations of bilingual language use that challenge traditional structural constraints of code-mixing \cite{kemaloglu2018patterns} and pose unique difficulties for otherwise simple foundational sequence-labeling tasks such as language identification (LID) and named entity recognition (NER).

\begin{table}[t]
\centering
\small
\begin{tabularx}{\linewidth}{@{}lX@{}}
\toprule
\textbf{Token} & \textbf{Gloss} \\
\midrule
\textit{influencerlarımız} &
\begin{tabular}[t]{@{}l@{}l@{}l@{}}
  \textbf{influencer}- & lar- & ımız\\[-1pt]
  influencer- & \textsc{pl}- & 1\textsc{pl.poss}
\end{tabular}\newline
`our influencers' \\[4pt]

\textit{frameworklerinin} &
\begin{tabular}[t]{@{}l@{}l@{}l@{}l@{}}
  \textbf{framework}- & ler- & i- & nin\\[-1pt]
  framework- & \textsc{pl}- & 3\textsc{pl.poss}- & \textsc{gen}
\end{tabular}\newline
`of their frameworks' \\[4pt]

\textit{tweetlememesi} &
\begin{tabular}[t]{@{}l@{}l@{}l@{}l@{}l@{}}
  \textbf{tweet}- & le- & me- & me- & si\\[-1pt]
  tweet- & \textsc{vblz}- & \textsc{neg}- & \textsc{nmlz}- & 3\textsc{sg.poss}
\end{tabular}\newline
`the fact that they do not tweet' \\
\bottomrule
\end{tabularx}
\caption{Mixed-language tokens with English stems and Turkish suffixes.}
\label{tab:examples-intro}
\end{table}

These linguistic challenges are compounded by a scarcity of annotated resources. To our knowledge, there is only one publicly available Turkish-English code-mixed text dataset, comprising 391 naturally occurring examples with word-level language labels \cite{yirmibesoglu-eryigit-2018-detecting}. We are not aware of any resources for evaluating NER for Turkish-English code-mixing, nor LID with mixed-language tokens. This paper fills that gap and makes the following contributions: 
\begin{itemize}[topsep=0pt,noitemsep]
    \item \textbf{TurEngMix}: a large-scale corpus of naturally occurring Turkish-English code-mixed social media posts, spanning a diverse range of online topics. Beyond this work, it supports future computational and sociolinguistic study of lexical innovation, code-mixing behavior, and multilingual language use.
    \item An annotated Turkish-English benchmark with token-level language and named-entity labels (15K tokens) derived from TurEngMix.
    \item A systematic evaluation of decoder LLM and fine-tuned encoder baselines for LID and NER on Turkish-English code-mixed text.
\end{itemize}

We release the corpus and annotated benchmark under a CC BY-NC-SA 4.0 license for non-commercial use. The corpus, benchmark, code, and preprocessing resources are publicly available. \footnote{https://huggingface.co/datasets/ilydoa/TurEngMix} \footnote{https://github.com/kemnguyenle/TurEngMix}

\section{Background}

\subsection{Turkish-English Code-Mixing}

Code-mixing is well-documented among Turkish speakers in both diaspora communities and online spaces. Turkish immigrant communities in Western Europe exhibit extensive code-mixing as a norm in everyday informal in-group conversation \cite{8d611477fb6f441b9bf392af0e345939}, and similar patterns have been observed among Turkish-English bilinguals in the United States, where intra-sentential switching occurs at particularly high rates \cite{koban2013intra}. Turkish-English code-mixing is also prevalent in social media contexts, where it serves a range of communicative functions including identity expression and tone-setting \cite{akgur2021case}.

Turkish-English written code-mixing is highly asymmetric. Turkish is agglutinative, verb-final, and suffixal. Following the Matrix Language Frame Model \cite{myers1993duelling}, this makes it natural for Turkish to provide the matrix frame and English to supply content stems that take Turkish inflection (e.g., English nouns and adjectives bearing Turkish case, possessive, or derivational suffixes). Mixed-language tokens (i.e., English roots with Turkish suffix chains) are frequent in informal written communication, such as in Turkish social media data \cite{yirmibesoglu-eryigit-2018-detecting}. Accurately processing such tokens is thus an essential prerequisite for downstream NLP modeling and analysis in this domain, but remains a challenge for extant systems. 

\subsection{Code-Mixing Resources and Evaluation}

Most existing code-mixed NLP resources are concentrated on a small set of language pairs, mainly Spanish-English and Hindi-English \cite{dogruoz-etal-2021-survey, winata2023decades}. Central benchmarks (e.g., LinCE \cite{aguilar-etal-2020-lince} and GLUECoS \cite{khanuja-etal-2020-gluecos}) explicitly acknowledge this skew as they provide evaluation limited to a few popular pairs. This imbalance is a core structural problem in the field \cite{dogruoz-etal-2023-representativeness}. More recent work has begun expanding corpus coverage to underrepresented language pairs. For example, VietMix \cite{tran-etal-2026-vietmix} introduced an expert-translated Vietnamese-English corpus for machine translation, demonstrating both the utility of carefully curated resources and the continued difficulty that modern models face on code-mixed text. Nevertheless, similar resources remain unavailable for many morphologically and typologically distinct language pairs, including Turkish-English. 

While performance on lower-level tasks such as LID and NER has improved for code-mixed data \cite{winata2023decades, pakray2025natural, sheth-etal-2025-comi}, these tasks remain challenging in realistic multilingual settings. Recent studies have shown that LID performance can degrade across domains \cite{goot-2025-identifying}, and code-switched LID remains difficult even across multiple language pairs, with existing approaches struggling to reliably identify language boundaries \cite{burchell-etal-2024-code}. Similarly, multilingual NER benchmarks continue to expose challenges involving noisy text, fine-grained entity categories, and complex entity structures \cite{malmasi-etal-2022-semeval}. Recent work has also explored specialized approaches for code-mixed LID, including training-free methods such as MaskLID \cite{kargaran-etal-2024-masklid}. However, there remains a gap between higher-resourced and lower-resourced language pairs. Moreover, performance on more complex NLP tasks such as sentiment analysis and question answering still lags behind that of non-mixed data. For example, CodeMixBench, a multilingual benchmark spanning eighteen languages (though no Turkic languages), reports persistent performance degradation on code-mixed inputs across a range of NLP tasks despite recent advances in LLMs \cite{yang2025codemixbench}.

There are some resources for tasks such as sentiment analysis and question-answering in Turkish \cite{ccoltekin2023resources}, and transformer models trained specifically for Turkish \cite{Schmitt_2026, uludogan-etal-2024-turna-custom}. For code-mixing, there exists a publicly available Turkish-German Twitter/X corpus \cite{cetinoglu-2016-turkish} and an audio corpus \cite{cetinoglu-2017-code}. However, we are only aware of one Turkish-English code-switched written corpus, which comprises 391 naturally-occurring samples collected from social media settings and is designed for the task of code-switching detection \cite{yirmibesoglu-eryigit-2018-detecting}. To our knowledge, TurEngMix is the first Turkish–English benchmark jointly annotated for named entity recognition and token-level language identification on naturally occurring forum text, and the largest such resource to include explicit metadata for English-origin stems with Turkish morphology.

\section{TurEngMix Corpus}

We first introduce the data collection and filtering steps to create the TurEngMix corpus and present corpus statistics. In the next section, we describe our annotated benchmark for LID and NER, followed by our evaluation of baseline models.

\subsection{Data Collection}

\begin{figure}[t]
\small
\fbox{
\parbox{\columnwidth}{
\vspace{0.5em}
\textbf{uı ux designer and front end developer} olarak uygun fiyatlı \textbf{b2c b2b}, wordpress, e-ticaret gibi çeşitli alanlarda \textbf{web siteleri} yapabilirim. ulaşmak isteyenler yeşillendirsin.

tanım : sertbest çalışan
\\[0.5em]
\textbf{Topic Tag:} freelance
}
}
\caption{Example \textit{Ekşi Sözlük} post collected for the TurEngMix corpus, with code-mixing (English words mixed with Turkish text) highlighted for reference.}
\label{fig:example}
\end{figure}

The corpus consists of public Turkish social media posts from the Turkish-language social media site \textit{Ekşi Sözlük}, a Reddit-like social media site with posts organized by topic tags, which function similarly to subreddits. The filtered dataset, token-level annotations, and preprocessing scripts are publicly available.

We collected 30K publicly-available posts using the \texttt{eksipy} API wrapper. Because posts could be collected by their topic tag category, we first curated a diverse list of topic tags that were likely to surface code-mixing, including tech topics, social media/entertainment, lifestyle, education, and more (see Appendix for the full topic list). The posts were retrieved as text strings. No usernames, profile information, timestamps, or other personally identifying metadata were retained.

\subsection{Data Filtering}

To increase the prevalence of Turkish-English code-mixing, we applied a three-stage filtering pipeline. First, the \texttt{langdetect} Python library was used to identify posts containing both Turkish and English. We then applied two GPT-4o filtering stages: the first performed a binary check for the presence of English lexical material, while the second repeated this while excluding posts in which apparent English content consisted solely of URLs, brand names, or other non-linguistic strings that could produce false positives. Lastly, posts containing fewer than three words or more than 500 words, where a word is defined as a whitespace-separated character sequence, were excluded after filtering.

To assess the filtering pipeline, we manually inspected 100 retained and 100 discarded posts. Of the retained posts, 71 contained Turkish-English code-mixing, compared with only 22 of the discarded posts, indicating that filtering substantially raised the density of code-mixed content relative to the unfiltered pool. Not every retained post is code-mixed, and some code-mixed posts are discarded. The retained set, therefore, is strongly enriched for code-mixing rather than exhaustive. The remaining monolingual posts serve as useful negative examples for downstream LID and NER analyses.

\subsection{Corpus Statistics}

\begin{table}[t]
\centering
\small
\begin{tabular}{lr}
\toprule
\textbf{Statistic} & \textbf{Value} \\
\midrule
Unique posts & 5,549 \\
Total tokens & 486,974 \\
Post length range & 3--500 words \\
Mean post length & 87.75 words \\
Median post length & 57 words \\
\bottomrule
\end{tabular}
\caption{Summary statistics of the TurEngMix corpus.}
\label{tab:dataset_stats}
\end{table}

\begin{figure}[h]
    \centering
    \includegraphics[width=1\linewidth]{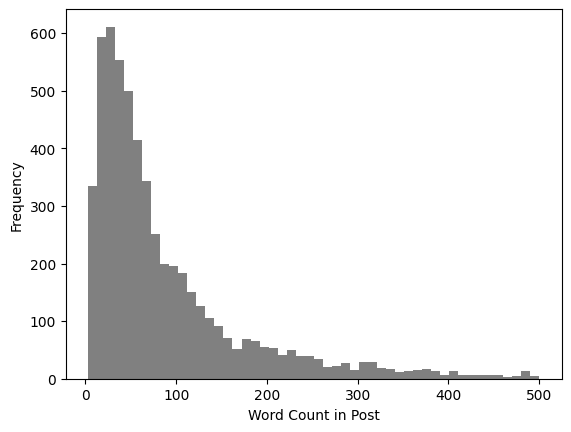}
    \caption{Word count distribution in the final dataset. Post word count was majority within 100 words, with longer posts scaling to 500.}
    \label{fig:length-dist}
\end{figure}

\begin{figure}[h]
    \centering
    \includegraphics[width=1\linewidth]{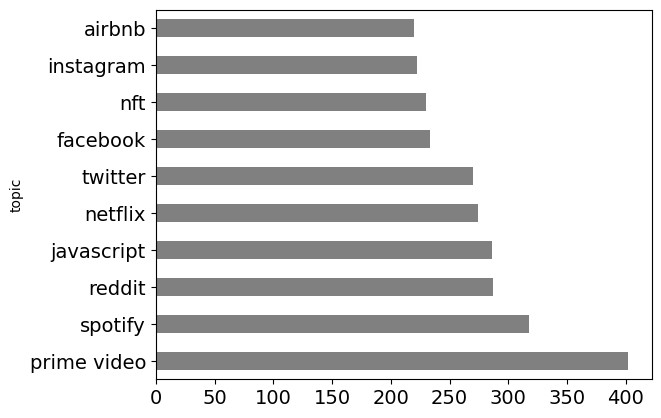}
    \caption{Ten most-frequent topic tags in the final dataset are primarily tech and entertainment.}
    \label{fig:topic-dist}
\end{figure}

The final TurEngMix Corpus consists of 5,549 unique posts comprising 486,974 tokens (Table~\ref{tab:dataset_stats}). Posts have a mean length of 87.75 words and a median length of 57 words. Figure \ref{fig:length-dist} shows the distribution of post lengths, illustrating that while most posts are relatively short, the corpus also captures longer-form text posts characteristic of online forums.

The corpus spans 71 unique topic tags. Technology and social media related topics (e.g., \textit{twitter}, \textit{netflix}, and \textit{javascript}) are among the most frequently represented categories  (Figure \ref{fig:topic-dist}). The prominence of technology and social media topics is consistent with the documented dominance of English in online and technical domains \cite{danet2007multilingual}, which makes such settings especially conducive to code-mixing \cite{androutsopoulos2013code, akgur2021case}.

\begin{table*}[t]
\centering
\small
\begin{tabularx}{\textwidth}{@{}lXX@{}}
\toprule
\textbf{Code-mixing type} & \textbf{Original post} & \textbf{English translation} \\
\midrule

Isolated English insertions &
her açtığımda belki gülhan şen kaydolmuştur diye \textbf{search} edip havamı aldığım \textbf{trendy} oluşum. &
The \textit{trendy} platform that, every time I open it, I \textit{search} thinking `maybe Gülhan Şen has signed up' and end up being disappointed. \\ 

\addlinespace

Mixed-language phrase &
\textbf{love is an excuse to get hurt} diye çok mantıklı bir açıklamasını yapmış birileri &
Someone gave a very reasonable explanation that \textbf{``love is an excuse to get hurt.''} \\

\addlinespace

Mixed-language token &
Bu \textbf{link'teki} video mutlaka izlenmeli bence. &
The video at \textbf{this link} should definitely be watched. \\

\bottomrule
\end{tabularx}
\caption{Different types of code-mixing found in the dataset, including isolated English insertions, longer mixed language phrases, and mixed-language tokens.}
\label{tab:examples}
\end{table*}

The corpus captures a broad spectrum of code-mixing behavior, including isolated English lexical insertions, longer mixed-language phrases, and morphologically integrated English stems with Turkish suffixes (Table~\ref{tab:examples}). As the TurEngMix corpus preserves naturally occurring Turkish-English discourse across diverse online contexts, it supports future investigations of code-mixing behavior, multilingual communication, and language variation, while also providing a resource for developing and evaluating NLP systems beyond LID and NER.

\section{LID and NER Annotated Benchmark}

\begin{figure*}[t]
    \centering
    \includegraphics[width=\textwidth]{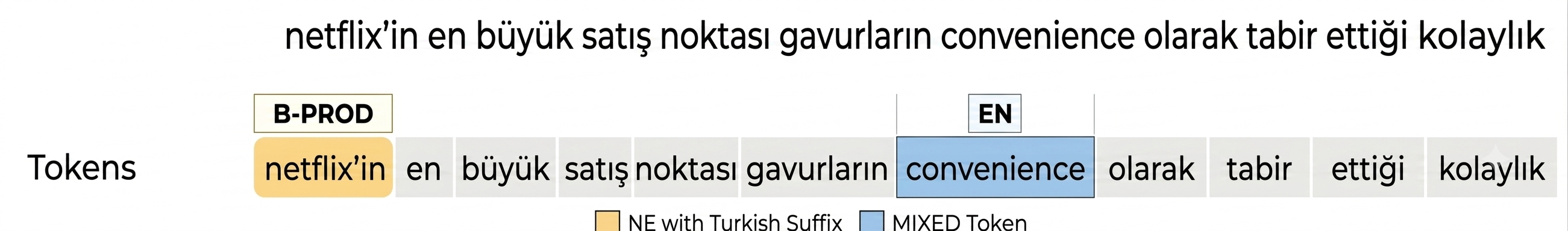}
    \caption{An example annotation sentence from the golden annotated dataset. Translation: \textit{Netflix's biggest selling point is the ease of use which the Westerners refer to as 'convenience.'}
    }
    
    \label{fig:sentence-annotation}
\end{figure*}

To enable standardized evaluation of foundational sequence labeling tasks on Turkish-English code-mixing, we constructed an annotated benchmark comprising 15,012 annotated tokens from the TurEngMix Corpus. By deriving the benchmark directly from naturally occurring social media posts, the annotated dataset preserves the linguistic characteristics of authentic Turkish-English code-mixed discourse.

\subsection{Benchmark Construction}
We first randomly sampled 250 posts from the corpus, then tokenized each post with a custom Python script that separates text based on whitespace and punctuation boundaries while preserving alphanumeric strings and morphologically mixed tokens. We removed hyperlinks, non-linguistic symbols, and special characters that did not contribute to language identification or named entity annotation. Punctuation marks such as apostrophes or quotation marks were retained because of their role in building morphologically complex words. The tokenization script and annotation guidelines are publicly available alongside the released dataset.

\subsubsection{Annotation Scheme}

Each token received a language identification label and a named entity label, as shown in Figure \ref{fig:sentence-annotation}.

\paragraph{LID}
We adopted the annotation framework of \citet{solorio-etal-2014-overview}, labeling each token as TR, EN, MIXED, OTHER, AMBIGUOUS, or NE. Following the shared task convention, all named entities were assigned the NE label regardless of language of origin or the presence of Turkish suffixes. For example, \textit{netflix'in} was labeled NE, not MIXED, despite containing a Turkish possessive suffix.

\paragraph{NER}
For NER, we followed the CALCS 2018 shared task guidelines \cite{aguilar-etal-2018-named}. Named entities were annotated as Person, Organization, Location, Group, Product, Title, Event, Time, or Other using the BIO tagging scheme.

\paragraph{Code-Mixing Metadata}
We also annotated for a broader "INTEGRATED" token category which captured whether an English-origin word, including named entities and brands, had a Turkish suffix. There were 386 tokens in this category, including tokens that had been labeled as NE in LID annotations but contained morphological mixing. This allowed us to understand the prevalence of the tokens that displayed the morphological mixing behavior. 

Posts were also labeled with the type of code-mixing present (see Table~\ref{tab:examples}), including INTEGRATED tokens embedded into the post, isolated English words inserted into Turkish text, and English phrases (> 2 words) present in the text. There was overlap between categories.

\subsubsection{Annotation Process}

The first author, fluent in both Turkish and English, annotated each token. To validate, inter-annotator agreement was calculated with a secondary annotator (also fluent in both languages) on a sample of 203 tokens from randomly sampled posts. A Cohen's Kappa score of 0.96 was achieved for LID and 0.94 for NER. Per-class agreement was high across categories (see Appendix~\ref{appendix:iaa}), with the lowest on MIXED ($\kappa$ = 0.75).

Disagreements primarily involved lexicalized English loanwords (e.g., \textit{videolar}, which may reasonably be interpreted as either Turkish or morphologically mixed) and distinctions between semantically related named entity categories (e.g., Product versus software-related terms). These cases were resolved through discussion, yielding complete agreement in the final annotations.

\subsection{Benchmark Statistics}

\begin{table}[t]
\centering
\small
\begin{tabular}{lr}
\toprule
\textbf{LID Label} & \textbf{Count} \\
\midrule
TR & 11,723 \\
EN & 1,992 \\
NE & 1042 \\
Mixed & 189 \\
Ambiguous & 62 \\
Other Language & 4 \\
\midrule
Total & 15,012 \\
\bottomrule
\end{tabular}
\caption{Distribution of LID labels. Note that NE tokens included additional morphologically mixed examples.}
\label{tab:lid_dist}
\end{table}

\begin{table}[t]
\centering
\small
\begin{tabular}{lr}
\toprule
\textbf{NER Label} & \textbf{Count} \\
\midrule
Non-Entity (O) & 13,971 \\
Product & 359 \\
Title & 246 \\
Organization & 129 \\
Person & 119 \\
Location & 88 \\
Other & 52 \\
Group & 26 \\
Event & 19 \\
Time & 3 \\
\midrule
\textbf{Total} & \textbf{15,012} \\
\bottomrule
\end{tabular}
\caption{Distribution of named entity labels.}
\label{tab:ner_distribution}
\end{table}

The benchmark exhibits substantial code-mixing, with a mean document-level Code-Mixing Index (CMI) of 13.25 (SD = 13.12) and 87.2\% of posts containing more than one language \cite{das-gamback-2014-identifying}. Although morphologically integrated tokens account for only 2.57\% of all annotated tokens, they appear in 66.0\% of posts, indicating that Turkish morphological integration is a frequent characteristic of code-mixed discourse despite its relatively low token frequency. Beyond morphologically integrated tokens, the benchmark also captures other common forms of Turkish-English code-mixing. A total of 154 posts contain isolated English insertions, while 148 contain longer embedded English phrases (Table~\ref{tab:pattern_statistics}).

\begin{table}[t]
\centering
\small
\begin{tabular}{lrr}
\toprule
\textbf{Split} & \textbf{Posts} \\
\midrule
Embedded Mixed Token & 165 \\
Isolated English Token & 154 \\
Embedded English Phrase & 148 \\
\bottomrule
\end{tabular}
\caption{Distribution of code-mixing patterns in the benchmark.}
\label{tab:pattern_statistics}
\end{table}

\begin{table}[t]
\centering
\small
\begin{tabular}{lccc}
\toprule
\textbf{Split} & \textbf{Posts} & \textbf{Sentences} & \textbf{Tokens} \\
\midrule
Train & 200 & 254 & 12,466 \\
Validation & 25 & 26 & 1,173 \\
Test & 25 & 41 & 1,373 \\
\midrule
Total & 250 & 321 & 15,012 \\
\bottomrule
\end{tabular}
\caption{Dataset statistics across train, validation, and test splits.}
\label{tab:split_statistics}
\end{table}

For the encoder models, the benchmark was partitioned into training, validation, and test splits using an 80/10/10 ratio, with posts (rather than individual tokens) assigned to splits to prevent information leakage across sets (Table~\ref{tab:split_statistics}). Each split maintained a similar distribution of LID and NER labels, with most labels represented (See Table~\ref{tab:lid_dist} and \ref{tab:ner_distribution} for benchmark label distributions). For NER, some of the rarest NE labels (Event, Time, etc) were not represented in the test set.

\section{Baseline Evaluations}

We present baseline evaluations of pretrained encoder and decoder models on two mature sequence-labeling tasks—LID and NER.

\subsection{Decoder Setup}

We selected GPT-4o and Qwen3-8B to represent two complementary classes of contemporary LLMs: a large proprietary model and a compact open-weight multilingual model.

For both tasks, models received complete posts reconstructed from the benchmark as sequences of pre-tokenized, indexed tokens, rather than receiving isolated words. This preserved the original sentential context of each social media post and allowed models to leverage context when predicting on ambiguous tokens. Models were prompted to assign a label to each pre-tokenized input token while preserving its provided token index. For LID, labels corresponded to the benchmark language categories. For NER, models predicted BIO-formatted named entity labels. Predicted labels were aligned with the benchmark using the provided token indices prior to scoring, and malformed outputs were treated as incorrect.

Model temperature was fixed at 0 to encourage deterministic outputs, and prompts required the model to assign exactly one label to each token. We evaluated both zero-shot and three-shot prompting to measure the effect of in-context examples. The three demonstration examples were selected to cover the range of language and named entity labels present in the benchmark. Complete prompts are included in the Appendix.

Predictions were produced for the complete benchmark (250 posts) and compared against the gold-standard annotations. We report macro-averaged F1 scores together with per-label F1 scores for both the LID and NER tasks.

\subsection{Encoder Setup}

To compare general-purpose LLMs with specialized encoder architectures, we evaluated three pretrained transformer encoders: XLM-RoBERTa \cite{conneau2020unsupervisedcrosslingualrepresentationlearning}, BERTurk \cite{yildirim2024finetuningtransformerbasedencoderturkish}, and TurkishBERTweet \cite{najafi2023turkishbertweetfastreliablelarge}. These models represent different sources of linguistic specialization: multilingual representation learning, general Turkish language modeling, and social-media-specific Turkish language modeling.

Models were fine-tuned for token classification using the Hugging Face Transformers framework on the TurEngMix Benchmark training split, with a majority-class baseline reported for comparison. Because encoder tokenizers operate on subword units, benchmark tokens were further segmented using each model's tokenizer, with labels assigned only to the first subword piece and continuation subwords ignored during training and evaluation.

Fine-tuning was performed using the Hugging Face Trainer with a learning rate of $2\times10^{-5}$, weight decay of 0.01, and a maximum of 10 epochs. Each model was trained using four random seeds. For each run, the checkpoint with the highest validation F1 score was evaluated on the held-out test split, and we report the mean and standard deviation across runs. Predictions were evaluated against the gold-standard annotations using macro-averaged and per-label F1 scores.

\section{Results}

\begin{table*}[htbp!]
\centering
\small
\begin{tabular}{lccccccc}
\toprule
\textbf{Model} 
& \textbf{Setting}
& \textbf{TR} 
& \textbf{EN} 
& \textbf{MIXED} 
& \textbf{NE} 
& \textbf{OTHER} 
& \textbf{AMBIGUOUS} \\
\midrule
GPT-4o & zero-shot 
& 0.93 & 0.89 & 0.42 & 0.62 & \textbf{0.19} & 0.00 \\
GPT-4o & three-shot 
& \textbf{0.98} & \textbf{0.90} & \textbf{0.45} & 0.67 & 0.13 & 0.00 \\
Qwen3-8B & zero-shot 
& \textbf{0.98} & 0.86 & 0.33 & 0.60 & 0.07 & \textbf{0.12} \\ 
Qwen3-8B & three-shot 
& \textbf{0.98} & 0.86 & 0.39  & 0.54 & 0.10 & 0.00 \\

\midrule

XLM-RoBERTa & fine-tuned 
& \textbf{0.98} & 0.87 & 0.00 & \textbf{0.68} & 0.00 & 0.00 \\
BERTurk & fine-tuned 
& 0.97 & 0.84 & 0.13 & 0.64 & 0.00 & 0.00 \\
TurkishBERTweet & fine-tuned 
& \textbf{0.98} & 0.84 & 0.22 & 0.56 & 0.00 & 0.00 \\

\bottomrule
\end{tabular}
\caption{Per-class F1 scores for LID. Decoder LLMs were tested on the full dataset. Encoder models were fine-tuned on the train/dev and evaluated on the test set.}
\label{tab:lid-per-class}
\end{table*}

\begin{table}[htbp!]
\centering
\small
\begin{tabular}{lcc}
\toprule
Model & Setting & Macro F1 \\
\midrule
GPT-4o & Zero-shot & 0.5170 \\
GPT-4o & Three-shot & \textbf{0.5238} \\
Qwen3-8B & Zero-shot & 0.4920 \\
Qwen3-8B & Three-shot & 0.4788 \\
\midrule
Baseline & -- & 0.1414 \\
XLM-RoBERTa & Fine-tuned & 0.4358 ± 0.0137 \\
BERTurk & Fine-tuned & \textbf{0.4734 ± 0.0355} \\
TurkishBERTweet & Fine-tuned & 0.4195 ± 0.0177 \\
\bottomrule
\end{tabular}
\caption{Macro F1 scores for LID task. Decoder LLMs were tested on the full dataset. Encoders were fine-tuned on the train/dev and evaluated on the test set. }
\label{tab:lid-main-results}
\end{table}

\subsection{Language Identification}

Distinguishing morphologically integrated English borrowings from ordinary Turkish or English tokens is the central difficulty in Turkish-English LID. Overall macro-F1 is modest across all models (Table~\ref{tab:lid-main-results}): GPT-4o is strongest, with three-shot prompting giving a small gain over zero-shot (0.5170$\rightarrow$0.5238), while Qwen3-8B does not benefit from demonstrations (0.4920$\rightarrow$0.4788). Fine-tuned encoders are comparable overall, with BERTurk and TurkishBERTweet leading among them. The aggregate scores, however, mask sharp per-class variation (Table~\ref{tab:lid-per-class}): every model labels Turkish and English tokens well (F1 > 0.84), but MIXED tokens and named entities are far harder. GPT-4o reaches only 0.45 F1 on MIXED tokens; the best encoder (TurkishBERTweet) reaches 0.22. Encoders match the LLMs on named entities (F1 0.56–0.68) but the results are much lower on MIXED, confirming that morphological integration—not language identity per se—is what current models struggle with (Table~\ref{tab:lid-per-class}).

\subsection{Named Entity Recognition}

Table~\ref{tab:ner-main-results} reports overall BIO-level NER performance. GPT-4o outperformed Qwen3-8B under both prompting settings, and three-shot prompting improved GPT-4o from a macro F1 of 0.5669 to 0.6140. In contrast, encoder models achieved considerably lower macro F1 despite high token-level accuracy, reflecting the strong class imbalance in the dataset where non-entity tokens account for approximately 93\% of all tokens.

\begin{table}[t]
\centering
\small
\begin{tabular}{lcc}
\toprule
Model & Setting & Macro F1 \\
\midrule
GPT-4o & Zero-shot & 0.5669 \\
GPT-4o & Three-shot & \textbf{0.6140} \\
Qwen3-8B & Zero-shot & 0.3993 \\
Qwen3-8B & Three-shot & 0.4162 \\
\midrule
Baseline & -- & 0.0502 \\
XLM-RoBERTa & Fine-tuned & 0.0772 $\pm$ 0.0246 \\
BERTurk & Fine-tuned & \textbf{0.1759 $\pm$ 0.0550} \\
TurkishBERTweet & Fine-tuned & 0.0821 $\pm$ 0.0273 \\
\bottomrule
\end{tabular}
\caption{BIO-level NER macro F1.}
\label{tab:ner-main-results}
\end{table}

\begin{table}[t]
\centering
\small
\begin{tabular}{llcc}
\toprule
\textbf{Model} 
& \textbf{Setting}
& \textbf{Non-NE}
& \textbf{NE} \\
\midrule
GPT-4o 
& Zero-shot 
& \textbf{0.98} 
& 0.78 \\

GPT-4o 
& Three-shot 
& \textbf{0.98} 
& \textbf{0.80} \\

Qwen3-8B 
& Zero-shot 
& 0.97 
& 0.70 \\

Qwen3-8B 
& Three-shot
& \textbf{0.98} 
& 0.73 \\

\midrule

XLM-RoBERTa 
& Fine-tuned 
& 0.97 
& \textbf{0.67} \\

BERTurk 
& Fine-tuned 
& 0.96 
& 0.47 \\

TurkishBERTweet 
& Fine-tuned 
& 0.97 
& 0.62 \\

\bottomrule
\end{tabular}
\caption{F1 scores for binary NER. BIO labels are collapsed into two categories: \textsc{NE} and \textsc{Non-NE}.}
\label{tab:ner-binary}
\end{table}

Because BIO evaluation jointly measures entity detection, entity classification, and boundary prediction, we additionally evaluate binary entity detection by collapsing all entity categories into entity (NE) and non-entity (Non-NE) (Table ~\ref{tab:ner-binary}). GPT-4o achieved an F1 of 0.80 on the NE class under three-shot prompting, suggesting that many BIO-level errors were associated with correctly identifying the specific entity type or boundary prediction rather than failure to detect named entities. Qwen achieved a lower F1 of 0.70 on the NE class, while encoder models ranged from 0.47 to 0.67.

Per-category evaluation reveals wide variation across entity types. Person, Location, and Title entities generally had higher F1 scores, whereas Organization, Other, Event and Time were substantially lower (Appendix Tables \ref{tab:ner-per-class-llm}-\ref{tab:ner-per-class-encoder}). Across all evaluated models, Organization and Product were the most frequently confused categories, particularly for technology-related entities whose interpretation depended on context (e.g., \textit{Google}, \textit{Netflix}, and \textit{Microsoft}).

\subsection{Error Analysis}

Qualitative inspection demonstrates consistent error patterns across both sequence labeling tasks.

\subsubsection{LID}

For language identification, both GPT-4o and Qwen frequently misclassified morphologically integrated English borrowings as ordinary Turkish tokens, particularly when multiple Turkish suffixes were attached (e.g., \textit{entryden}, \textit{influencerlarımız}, and \textit{editleyip}). Named entities containing Turkish suffixes (e.g., \textit{Microsoft'un} and \textit{Sony'nin}) were also commonly predicted as \textsc{mixed} rather than \textsc{NE}, suggesting that models relied heavily on surface morphology instead of treating named entities as a separate language category. Conversely, English compounds containing hyphens (e.g., \textit{k-food} and \textit{marketing-mix}) were often incorrectly labeled as \textsc{mixed} despite containing no Turkish morphology.

\subsubsection{NER}

For named entity recognition, the most common errors involved morphologically integrated English borrowings and technology-related named entities. Technology entities frequently produced Product-Organization confusion (e.g., \textit{Twitter}, \textit{Google}, \textit{Cloudflare}, and \textit{Last.fm}), while morphologically integrated entities such as \textit{reactlerle} and \textit{python'daki} were often missed after Turkish suffix attachment.

\subsubsection{Morphological Integration in Named Entities}

To further examine whether morphological integration affects NER performance, we conducted an error analysis on the GPT-4o three-shot and Qwen three-shot settings. We compared NER error rates between tokens containing English-origin stems bearing Turkish suffixes (including morphologically adapted named entities) and all other evaluated tokens (Table~\ref{tab:borrowed-errors}). 
GPT-4o produced errors on 19.7\% of mixed tokens compared with only 3.8\% of other tokens. A chi-square test confirmed that this difference was highly significant ($\chi^2=231.83$, $p<0.001$), corresponding to an odds ratio of 6.21. Qwen exhibited an even larger disparity, with error rates of 33.5\% and 5.3\%, respectively ($\chi^2=507.17$, $p<0.001$; odds ratio = 8.84). These results suggest that morphological integration introduces an additional challenge beyond general code-mixed language processing.

\begin{table}[t]
\centering
\small
\begin{tabular}{lcc}
\toprule
Model & Other Tokens & Mixed Tokens \\
\midrule
GPT-4o & 3.8\% & 19.7\% \\
Qwen3-8B & 5.3\% & 33.5\% \\
\bottomrule
\end{tabular}
\caption{Comparison of NER error rates between morphologically integrated English borrowings and other evaluated tokens.}
\label{tab:borrowed-errors}
\end{table}

\section{Conclusion}

This work addresses gaps in Turkish-English code-mixed NLP by introducing a large-scale corpus, an expert-annotated benchmark for language identification and named entity recognition, and systematic evaluations of current LLM and encoder-based models. Our analysis reveals that morphologically mixed tokens remain a persistent challenge for multilingual NLP systems, highlighting an underexplored difficulty for agglutinative languages in code-mixed settings. These findings suggest that progress on code-mixed NLP for agglutinative languages may require morphology-aware approaches, such as sub-token language attribution or morphological segmentation as prepocessing, rather than scale alone. By releasing these resources and evaluations, we aim to support future research on multilingual language understanding and improve the representation of diverse language varieties in NLP.

\section{Limitations}

As the data was collected from \textit{Ekşi Sözlük}, a social media site with a specific user base, this Turkish-English code-mixing resource is thus only most accurately representative of the code-mixing practiced by that online community, and does not account for different patterns in other domains. Furthermore, because topic tags likely to contain Turkish-English code-mixing were prioritized during corpus construction, the corpus is not intended to estimate the prevalence of code-mixing in Turkish online communication. As the site is public, pretraining contamination is possible. Future work could expand to other online communities or other settings (e.g. spoken, in-person). 

A further consideration is that GPT-4o served both as a filtering component during corpus construction (§3.2) and as an evaluated baseline. Posts in the corpus are, by construction, posts in which GPT-4o detected English lexical material, which may favor GPT-4o's LID performance relative to models that played no role in dataset creation. We partially mitigate this by including a langdetect-based first stage and by manual inspection of retained and discarded posts, but results for GPT-4o should be interpreted with this selection effect.

The annotation scheme follows established task definitions, but the separate NE category can obscure cases where named entities also exhibit Turkish morphological integration; we partially address this by separately annotating the "INTEGRATED" category for all English-origin tokens with Turkish morphological suffixes. The LLM prompts also did not systematically provide demonstrations for every annotation category, leaving the effect of alternative prompting strategies as an important direction for future work.

Another limitation was the class imbalance present in the dataset. Some LID and NER categories contained only a small number of instances, limiting the reliability of per-category evaluation for rare labels. For LID, the majority of tokens were TR tokens. For NER, O (non-entity) tokens comprised the majority of the dataset. 

\bibliography{anthology, custom}

\newpage
\appendix
\renewcommand{\thefigure}{A\arabic{figure}}
\renewcommand{\thetable}{A\arabic{table}}
\setcounter{figure}{0}
\setcounter{table}{0}

\section{Appendix}

\subsection{Dataset Collection}

The TurEngMix Corpus was collected from publicly available posts on
\textit{Ekşi Sözlük} during November 2025. Candidate posts were retrieved
using topic tags selected to increase the likelihood of Turkish-English
code-mixing.

\subsubsection{Topic Tags}

\begin{lstlisting}[language=Python]
topic_names = [
    #Tech / Programming / AI
    "spotify", "google", "python", "ai", "chatgpt", "openai", "midjourney", "dalle", 
    "javascript", "react", "nodejs", "git", "blockchain", "web3", "crypto", "nft", "startup", "venture capital",

    #Social Media / Internet Culture
    "facebook", "instagram", "twitter", "youtube", "tiktok", "linkedin", "discord", "reddit", 
    "influencer", "meme", "shitpost", "gaming", "twitch", "streamer",

    #Entertainment / Gaming / Anime
    "netflix", "prime video", "disney+", "minecraft", "lol", "anime", "kpop", "manga", "webtoon", 
    "rap", "hiphop", "edm", "dj",

    #Sports
    "futbol", "basketbol", "tennis", "formula 1", "voleybol", "gym", "workout", "fitness",

    #Lifestyle / Travel / Food / Shopping
    "yemek", "tatil", "seyahat", "alışveriş", "fashion", "sneakers", "nike", "adidas", 
    "apple", "samsung", "mobile", "app", "airbnb", "hotel", "vegan", "fitness lifestyle",

    #Daily Life / Work / Education
    "üniversite", "iş", "freelance", "marketing", "growth", "seo", "branding", "startup life", 
    "gündem", "aşk", "sağlık", "covid", "ekonomi", "politika", "haberler", "finance",

    #Misc / Fun / Pop Culture
    "lol memes", "viral", "challenge", "trend", "gaming memes", "internet slang", 
    "funny videos", "edutainment", "tech news", "apps review", "music", "concert", "festival"
]
\end{lstlisting}

\subsection{Corpus Filtering}

Corpus construction employed a multi-stage filtering pipeline to identify
posts containing Turkish-English code-mixing while removing noise.

\subsubsection{Stage 1 Prompt: English Word Detection}

Initial filtering prompt used to identify posts containing English words.

\begin{lstlisting}[language=Python]
prompt_text = """
Detect if the given Turkish social media post contains any English words. If English words are present, return true.
      
Steps
1. Language Detection: Detect whether the given text contains English words.
      
Output Format:
Detection Result: Output "True" if English words are present, otherwise "False". Number your answers to match the post numbers, and put each response on its own line.

       Example output:
       1. True
       2. False
       etc
"""
\end{lstlisting}

\subsubsection{Stage 2 Prompt: Code-Mixing Verification}

Second-stage filtering prompt providing a more explicit instructions for detection of Turkish-English code-mixing.

\begin{lstlisting}[language=Python]
prompt_text = """
You will be given Turkish social media posts. Go through the entire post, word by word, and detect if it contains any English words mixed into the Turkish, including individual English words or sentences/phrases in English. Don't count URLs or brand names as English words.

If English words are present in the post, return true.

Steps
1. Language Detection: Detect whether the given post contains English words.

Output Format:
Detection Result: Output "True" if English words are present, otherwise "False".
Number your answers to match the post numbers, and put each response on its own line.

       Example output:
       1. True
       2. False
       etc
"""
\end{lstlisting}


\subsubsection{Language Identification Prompt}

Prompt template used for LLM-based language identification experiments.

\begin{lstlisting}[language=Python]
prompt_text = """
Determine the language label for each token in the input text.

Use the following labels:
TR: Turkish
EN: English
MIXED: Mixed Turkish and English
OTHER: Any other language
NE: Named entity (representing all or part of a named entity)
AMBIGUOUS: Could belong to either language in context

The input consists of numbered tokens. Each input line contains an integer index, followed by a tab character, followed by a token.

For every input line:
- Output exactly one line.
- Preserve the same INDEX.
- Assign exactly one language label.
- Do not skip any indices.
- Do not duplicate any indices.
- Do not renumber the indices.

Output each line in the following format:

INDEX|LABEL

The number of output lines must exactly equal the number of input lines.

Return only the INDEX|LABEL pairs, one per line. Do not include the original tokens, explanations, numbering, headings, blank lines, or any other text.
"""
\end{lstlisting}

\subsubsection{Named Entity Recognition Prompt}

Prompt template used for LLM-based named entity recognition experiments.

\begin{lstlisting}[language=Python]
prompt_text = """
Analyze the following text and identify named entities using the BIO tagging scheme.

Use the following BIO format:
- B-: The first token of a named entity.
- I-: Any subsequent token belonging to the same named entity.
- O: A token that is not part of any named entity.

Use the following entity categories:

- B-PER / I-PER (Person): Proper names or nicknames that uniquely identify a real or fictional person, including artists and famous people.
- B-ORG / I-ORG (Organization): Companies, institutions, corporations, and other entities that have members or employees and act as a whole. When an entity could refer to either an organization or another category (e.g., Facebook), use the surrounding context to determine the correct label.
- B-LOC / I-LOC (Location): Physical places that people can visit, including countries, cities, addresses, facilities, landmarks, restaurants, and tourist attractions. When an organization name is used to refer to a physical place, label it as LOC.
- B-GROUP / I-GROUP (Group): Sports teams, music bands, duos, and similar groups of people.
- B-PROD / I-PROD (Product): Manufactured products, medicines, food products, well-defined services, website applications, website accounts, devices, and similar products. Distinguish products from organizations using context.
- B-TITLE / I-TITLE (Title): Titles of creative works such as movies, books, TV shows, songs, and similar media.
- B-EVENT / I-EVENT (Event): Named events such as concerts, competitions, conferences, festivals, and award ceremonies. Do not use this category for holidays.
- B-TIME / I-TIME (Time): Months, weekdays, seasons, holidays, and recurring calendar dates. Do not label hours, minutes, seconds, or relative temporal expressions such as "yesterday," "tomorrow," "week," or "year."
- B-OTHER / I-OTHER (Other): Named entities that do not fit the categories above, including nationalities, languages, music genres, and similar entities.
- O (None): Not part of a named entity

The input consists of numbered tokens. Each input line contains an integer index, followed by a tab character, followed by a token.

For every input line:
- Output exactly one line.
- Preserve the same INDEX.
- Assign exactly one named entity label.
- Do not skip any indices.
- Do not duplicate any indices.
- Do not renumber the indices.

Output each line in the following format:

INDEX|LABEL

The separator between INDEX and LABEL must be the vertical bar character "|".
Do not use tabs.
Do not use spaces.
Correct:
1|B-PER
Incorrect:
1 B-PER
1    B-PER
1: B-PER

The number of output lines must exactly equal the number of input lines.

Return only the INDEX|LABEL pairs, one per line. Do not include the original tokens, explanations, numbering, headings, blank lines, or any other text.
"""
\end{lstlisting}

\subsection{Per-Class Inter-Annotator Agreement}
\label{appendix:iaa}

Table~\ref{tab:lid-iaa-per-class} displays inter-annotator agreement for LID per class. Table~\ref{tab:ner-iaa-per-class} displays inter-annotator agreement for NER per class. All disagreements were resolved. 

\begin{table}[h]
\centering
\small
\begin{tabular}{lcc}
\toprule
\textbf{LID Label} & \textbf{Cohen's $\kappa$} \\
\midrule
TR & 0.98 \\
EN & 0.98 \\
NE & 0.94 \\
MIXED & 0.75 \\
\bottomrule
\end{tabular}
\caption{Per-class inter-annotator agreement for language identification (LID).}
\label{tab:lid-iaa-per-class}
\end{table}

\begin{table}[h]
\centering
\small
\begin{tabular}{lcc}
\toprule
\textbf{NER Label} & \textbf{Cohen's $\kappa$} \\
\midrule
B-LOC & 1.00 \\
I-LOC & 1.00 \\
B-PER & 1.00 \\
I-PER & 1.00 \\
B-PROD & 1.00 \\
B-TITLE & 1.00 \\
I-TITLE & 1.00 \\
O & 0.94 \\
B-OTHER & 0.00 \\
\bottomrule
\end{tabular}
\caption{Per-class inter-annotator agreement for named entity recognition (NER). The low $\kappa$ score for B-OTHER is likely due to the small number of instances in this category.}
\label{tab:ner-iaa-per-class}
\end{table}

\subsection{Additional Experimental Results}

\subsubsection{Per-Class Named Entity Recognition}

Table~\ref{tab:ner-per-class-llm} presents per-class named entity recognition results for GPT4o and Qwen. 

Table~\ref{tab:ner-per-class-encoder} presents per-class named entity recognition for the encoder models.

\begin{table*}[h]
\centering
\small
\begin{tabular}{lcccccccccc}
\toprule
\textbf{Model} &
\textbf{O} &
\textbf{ORG} &
\textbf{PER} &
\textbf{LOC} &
\textbf{PROD} &
\textbf{GROUP} &
\textbf{TITLE} &
\textbf{EVENT} &
\textbf{TIME} &
\textbf{OTHER} \\
\midrule
GPT-4o Zero-shot
& 0.98
& 0.48
& 0.79
& 0.83
& 0.53
& 0.35
& 0.69
& 0.46
& 0.20
& 0.29 \\

GPT-4o Few-shot
& 0.98
& \textbf{0.57}
& \textbf{0.83}
& \textbf{0.84}
& \textbf{0.67}
& \textbf{0.42}
& \textbf{0.72}
& \textbf{0.43}
& 0.09
& 0.34 \\

Qwen3-8B Zero-shot
& 0.97
& 0.43
& 0.55
& 0.65
& 0.46
& 0.35
& 0.66
& 0.00
& 0.09
& \textbf{0.43} \\

Qwen3-8B Few-shot
& 0.98
& 0.47
& 0.63
& 0.64
& 0.52
& 0.28
& 0.53
& 0.00
& 0.07
& 0.34 \\
\bottomrule
\end{tabular}
\caption{Per-class F1 scores for decoder LLMs on NER task. O denotes non-entity tokens.}
\label{tab:ner-per-class-llm}
\end{table*}

\begin{table*}[h]
\centering
\small
\begin{tabular}{lccccccc}
\toprule
\textbf{Model} &
\textbf{O} &
\textbf{LOC} &
\textbf{ORG} &
\textbf{PER} &
\textbf{PROD} &
\textbf{TITLE} &
\textbf{OTHER} \\
\midrule
Baseline
& 1.00
& 0.00
& 0.00
& 0.00
& 0.00
& 0.00
& 0.00 \\

XLM-RoBERTa
& 0.97
& 0.00
& 0.00
& 0.00
& 0.28
& 0.00
& 0.00 \\

BERTurk
& 0.97
& 0.00
& 0.091
& 0.44
& 0.27
& 0.14
& 0.00 \\

TurkishBERTweet
& 0.96
& 0.00
& 0.00
& 0.00
& 0.27
& 0.06
& 0.00 \\
\bottomrule
\end{tabular}
\caption{Per-class F1 scores for named entity recognition using encoder models. O denotes non-entity tokens.}
\label{tab:ner-per-class-encoder}
\end{table*}

\subsection{Error Analysis Examples}

Table~\ref{tab:error-analysis-examples} presents representative model errors
involving Turkish morphological integration and named entity ambiguity.

\begin{table*}[t]
\centering
\small
\begin{tabular}{p{0.1\linewidth} p{0.1\linewidth} p{0.15\linewidth} p{0.6\linewidth}}
\toprule
\textbf{Token} & \textbf{Gold Label} & \textbf{Model Prediction} & \textbf{Error Explanation} \\
\midrule

\textit{influencerlar} 
& MIXED 
& TR 
& English-origin stem \textit{influencer} with Turkish plural/possessive morphology. The token requires identifying both the English lexical root and Turkish morphological integration. \\\midrule

\textit{google'a}
& B-PROD 
& B-ORG 
& In context, \textit{Google} refers to the search product/service rather than the organization. \\\midrule

\textit{sony'nin}
& NE 
& MIXED 
& The token contains a Turkish possessive suffix attached to a named entity. The model incorrectly prioritized morphological mixing over named entity identification. \\

\bottomrule
\end{tabular}
\caption{Representative error cases involving morphological integration and named entity ambiguity.}
\label{tab:error-analysis-examples}
\end{table*}


\subsection{Released Resources}

The following resources are publicly available:

\begin{itemize}
    \item TurEngMix Corpus
    \item TurEngMix Benchmark
    \item Annotation guidelines
    \item Dataset construction scripts
    \item Training and evaluation scripts
    \item LLM prompting scripts
    \item Fine-tuning code
\end{itemize}

\end{document}